# An Approach for Reducing Outliers of Non Local Means Image Denoising Filter

Raka Kundu*, Amlan Chakrabarti and Prasanna Lenka

**ABSTRACT**—We propose an adaptive approach for 'non local means (NLM)' image filtering termed as 'non local adaptive clipped means (NLACM)', which reduces the effect of outliers and improves the denoising quality as compared to traditional NLM. Common method to neglect outliers from a data population is computation of mean in a range defined by mean and standard deviation. In NLACM we perform the median within the defined range based on statistical estimation of the neighborhood region of a pixel to be denoised. As parameters of the range are independent of any additional input and is based on local intensity values, hence the approach is adaptive. Experimental results for NLACM show better estimation of true intensity from noisy neighborhood observation as compared to NLM at high noise levels. We have verified the technique for speckle noise reduction and we have tested it on ultrasound (US) image of lumbar spine. These ultrasound images act as guidance for injection therapy for treatment of lumbar radiculopathy. We believe that the proposed approach for image denoising is first of its kind and its efficiency can be well justified as it shows better performance in image restoration.

Index Terms—Image denoising, non local means, sigma clipped Euclidean median, central tendency, edge preserving, ultrasound image.

## I. INTRODUCTION

LOWER back pain is a common condition which most people experience at some stage of their life [1]. Beside physical therapy, injection therapy that is targeted to nerve roots or facet joints is well established in the treatment of lumbar radiculopathy. Injection therapy has been done without any image guidance in the past years. Nowadays imaging guided techniques has made injection procedure easy and has increased the success rate of pain management. US have proved to be sufficiently reliable and accurate in the demonstration of lumbar paravertebral anatomy [1, 2]. US are highly sensitive to speckle noise and many researches [3] have been performed for elimination of noise from US images.

NLM denoising is performed as weighted mean of all other pixels of the noisy image where pixels with similar neighborhoods are assigned higher weights and vice versa. Considering $V_i$ as ith pixel of noisy image and $V'_i$ as the corresponding pixel of denoised image, we compute $V'_i$ as

$$V'_i = \frac{\sum_j w_{ij} V_j}{\sum_j w_{ij}} \qquad (1)$$

Where, $w_{ij}$ is the weight between pixels *i* and *j*, $V_j$ symbolize other pixels of the noisy image. In general, the search domain of j is lessened to a local domain of dimension *s x s*, which is concentrated around present pixel *i* [4], that is intended for denoising. Hence $V'_i$ (*equation 1*) is reformed to

$$V'_i = \frac{\sum_{j \epsilon s_i} w_{ij} V_j}{\sum_{j \epsilon s_i} w_{ij}} \quad (2)$$

Another idea for weight computation in [4] is to replace individual pixels *i* and *j* by image patches (*P*), where $P_i$ and $P_j$ are centered at *i* and *j* respectively. So, the revised weighted equation is:

$$w_{ij} = \exp(-\frac{||P_i - P_j||^2}{h^2}) \quad (3)$$

*||P||* is the Euclidean norm of *P*, *r x r* is size of *P*, and *h* [4] is denoising manipulation parameter. NLM was proposed by Buades et al [4, 5] in 2005. Many research works has been done on acceleration of NLM algorithm [6, 7, 8, 9]. Also work has been performed on quality enhancement of NLM [10, 11, 12, 13].

Mean is the common measure of central tendency to estimate the denoised amplitude value of a pixel from its local neighborhood. Its realization is best when data population is large enough and is devoid of outliers. In this paper, we discuss an algorithm for NLM that selects pixels of significant intensity value for computation of weighted Euclidean mean. A range is considered using standard deviation and mean for denoising by NLM. This is named as 'non local sigma clipped Euclidean means (NLSCEM)'. Also a modification of NLSCEM range is performed by replacing mean with median. The approach is named as 'non local adaptive clipped means (NLACM)'. The technique effectively reduced the affect of outliers and improved the denoising quality in terms of noise cleaning and edge preservation of the image.

## II. NON LOCAL SIGMA CLIPPED EUCLIDEAN MEANS

Noise (outliers) makes statistical mean less efficient for the calculation of central tendency. It is observed that defining a range with use of mean and standard deviation (measure of dispersion) and then calculation of weighted Euclidean mean (for NLM) within that range reduces outlier's affect at high noise levels. Let, **p** is mean of a data set *p* (elements in *s x s*). The standard deviation of *p* is expressed as:

$$sd = \sqrt{\frac{\sum_{i=1}^{N}(p_i - \mathbf{p})^2}{N}} \quad (4)$$

where *N* is number of elements in *p*. The upper and lower limits within which Euclidean mean is to be calculated from *p* is given by

$$p_l = \mathbf{p} - sd \quad (5)$$

$$p_u = \mathbf{p} + sd \quad (6)$$

The choices of upper and lower limits are based on the standard deviation and mean of the local neighborhood intensities. The parameters of the limits ($p_l$ and $p_u$) are computed adaptively and is not dependent on any additional input to the filter. NLSCEM is conveyed in Algorithm 1. Step II of the algorithm engage calculation of sigma clipped Euclidean mean for NLM. A sum of all the amplitude values of patch ($P_j$) is taken into account rather than considering a single pixel amplitude value $V_j$. So, every patch $P_j$ is presented by single value $p_j$ where, $p_j$ are elements of p within s x s. Based on elements of p mean (*p*) and standard deviation (*sd*) is calculated. The amplitude range within which pixels are to be considered for computation of weighted Euclidean mean is given by $p_l$ and $p_u$.

**Algorithm 1 Non Local Sigma Clipped Euclidean Means**

**INPUT:** Noisy image $V$, parameters $s, r, h$. $s$: Local neighborhood size, $r$: Patch size, $h$: Control parameter of filter.
**OUTPUT:** Denoised image $V'$.

Step I: Extract patch $P_i$ of size $r \times r$ for every pixel $i$ and centered at $i$.
Step II: For every pixel $i$ of $V$, perform:
(a) Set $wij = exp(-||Pi - Pj||^2/h^2)$, where $j \epsilon s_i$.
(b) Calculate $p_j$ = sum $(Pj)$, where $j \epsilon s_i$ and $p$ symbolize the new intensity set of the local neighborhood.
(c) Select clipped data set $p'$ of $p$.
(i) Calculate **p** = mean $(p)$.
(ii) Calculate $sd$ = {standardDeviation $(p)$}.
(iii) Calculate upper limit $p_u$ and the lower limit $p_l$ of $p$ using **p** and $sd$ (eqn. 5 and eqn. 6).
(iv) Let $p'_k$ be the $k^{th}$ element of the clipped set $p'$.
Form the clipped data set $p'$ where, every element $p'_k \epsilon p$ and satisfies the constraint $(p'_k \geq p_l)$ & $(p'_k \leq p_u)$.
(d) Calculate weighted mean of the clipped set $p'$
$P'_i = \sum_{j \epsilon p'} w_{ij} P_j / \sum_{j \epsilon p'} w_{ij}$
(e) Assign center pixel of $P'_i$ as denoised value of $V'_i$.

The equation for NLSCEM is as follows

$$P'_i = \frac{\sum_{j \epsilon p'} w_{ij} P_j}{\sum_{j \epsilon p'} w_{ij}} \quad (7)$$

where $P_j$ represent patches within $p_u$ - $p_l$ and $w_{ij}$ represent associated weights within the same clipped domain. Eventually the center pixel of $P'_i$ gives the denoised value of NLSCEM.

### III. NON LOCAL ADAPTIVE CLIPPED MEANS

Median is robust to outliers than mean [15, 12]. It raised a question that what will happen if we substitute mean by median in the defined range of NLSCEM. This gave a new observation where, this substitution geared up the performance of NLSCEM and diminished affect of outliers. Following are the revised equations of NLSCEM where median $(p)$ stands for median value of $p$.

$$sd = \sqrt{\frac{\sum_{i=1}^{N}(p_i - median(p))^2}{N}} \quad (8)$$

$$p_l = median(p) - sd \quad (9)$$

$$p_u = median(p) + sd \quad (10)$$

Parameters of value *s=10, r=3* and *h=10 x noise variance x 100/255* were set for all experiments of this paper. A better and different parameter tuning can be approached for the filter. In Fig. 1 an experiment is performed on *Checker* image at *noise variance=0.08*. The plot shows a clean edge and the respective edges obtained after denoising by NLM, NLSCEM and NLACM. It can be noticed that the edge obtained after denoising by NLACM is near to the true edge in comparison to standard NLM and NLSCEM. A zoomed portion of the Checker image, its noisy version and its respective denoised images are shown in Fig. 2 for visual comparison.

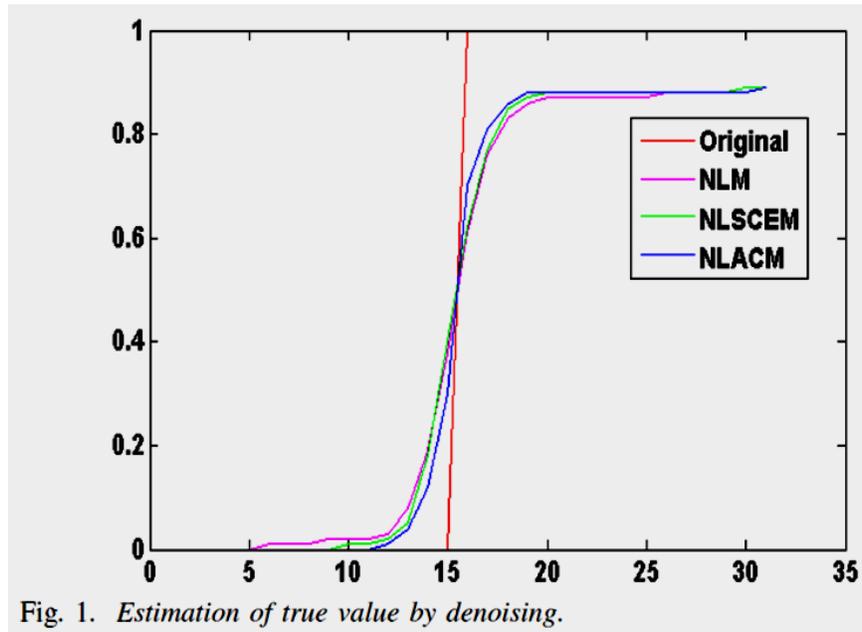

Fig. 1. *Estimation of true value by denoising.*

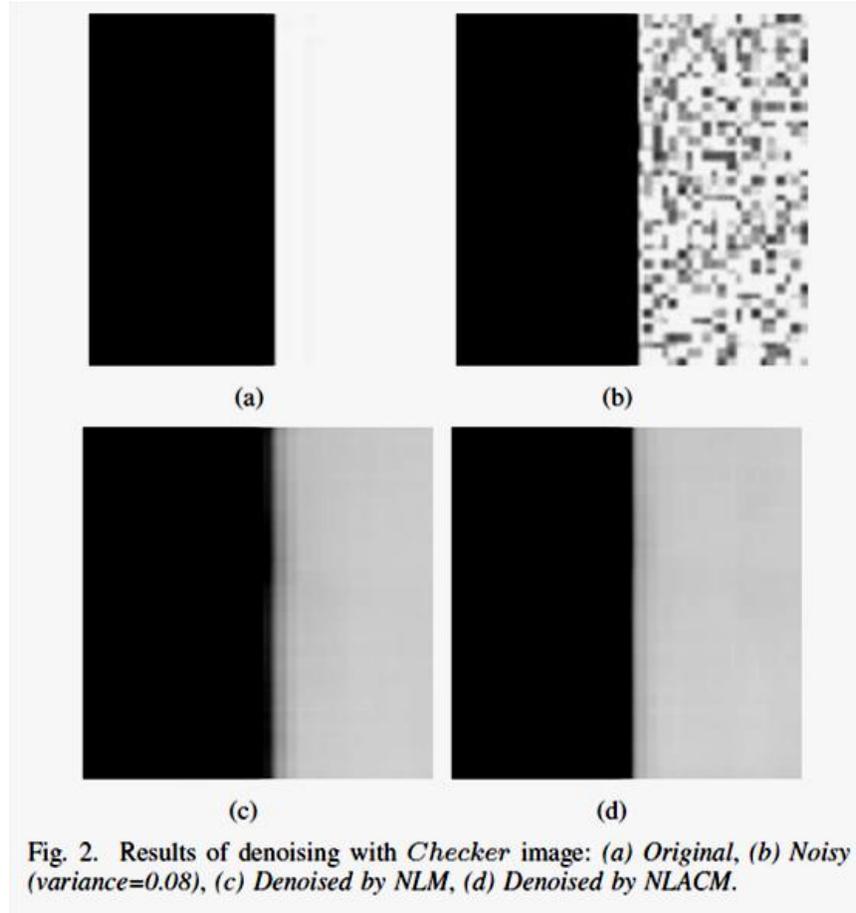

Fig. 2. Results of denoising with *Checker* image: *(a) Original, (b) Noisy (variance=0.08), (c) Denoised by NLM, (d) Denoised by NLACM.*

**IV. EXPERIMENTS**

The 'peak signal to noise ratio (PSNR)'data is accounted in Table 1. Speckle noise was introduced to the test image. Our technique shows higher PSNR values than standard NLM at higher noise levels. In the final experiment (Fig. 3) we concentrated on visual comparison of US lumbar image (taken from reference 2), at introduced speckle *noise variance 0.1*. The related PSNR metric values obtained from NLM and NLACM denoising techniques are *22.88 dB* and *23.16 dB* respectively.

TABLE I
PSNR DATA OF NLM, NLSCEM AND NLACM.

| | Method | 0.01 | 0.02 | 0.03 | 0.04 | 0.05 | 0.06 | 0.07 | 0.08 | 0.09 | 0.1 |
|---|---|---|---|---|---|---|---|---|---|---|---|
| PSNR (dB) | | | | | | | | | | | |
| Checker | NLM | 30.38 | 27.32 | 25.30 | 23.25 | 21.60 | 20.26 | 19.01 | 17.90 | 17.05 | 16.12 |
| | NLSCEM | 30.33 | 27.31 | 25.29 | 23.26 | 21.59 | 20.42 | 19.50 | 18.76 | 18.32 | 17.80 |
| | NLACM | 30.36 | 27.32 | **25.31** | **23.28** | **21.65** | **20.56** | **19.71** | **19.01** | **18.54** | **18.01** |

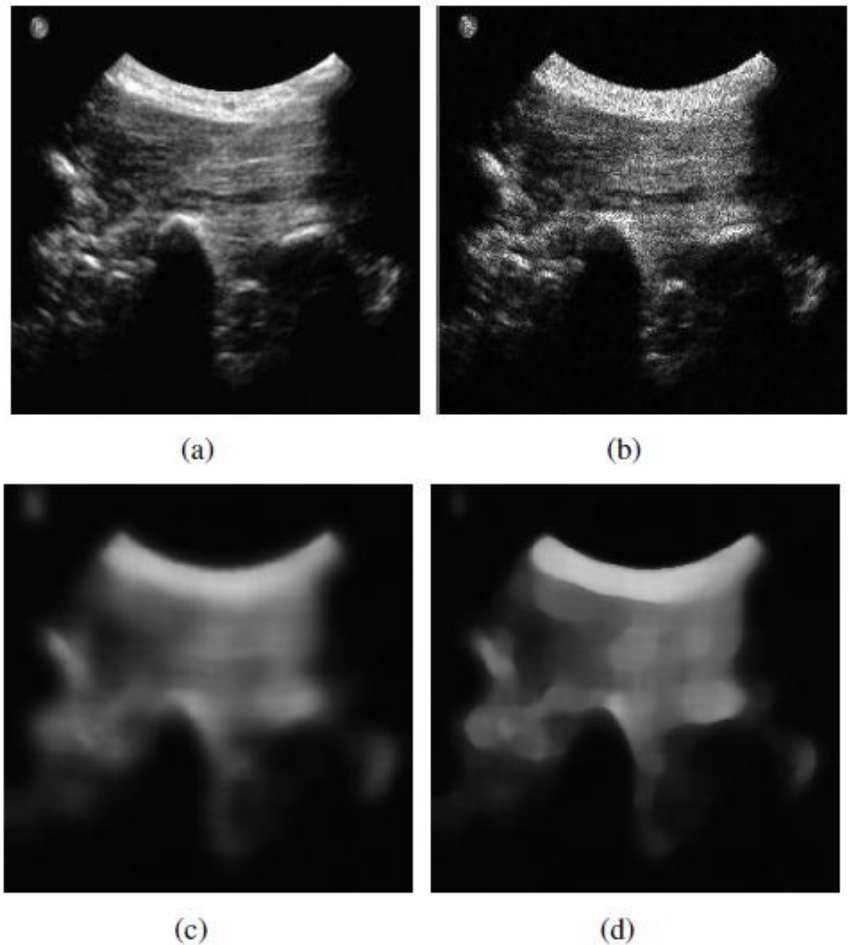

Fig. 3. Results of denoising for *Ultrasound* image: *(a) Original, (b) Noisy (variance=0.1), (c) Denoised by NLM, (d) Denoised by NLACM.*

## V. CONCLUSION

NLACM enhances the robustness of NLM algorithm with an ability to better restore image edges and remove noise effectively from US image and other images. The algorithm discards outliers by defining a range adaptively. This simple but better amplitude estimator will also profit other image denoising techniques.

This research was supported by Center of Excellence in Systems Biology and Biomedical Engineering (TEQIP PHASE-II), University of Calcutta and National Institute for the Orthopaedically Handicapped, Kolkata, India.

***R. Kundu** is Ph. D Fellow (SRA) and with Biomedical Image Processing and Analysis research at the Department of A. K. Choudhury School of Information Technology, University of Calcutta, Kolkata, 700009, India (e-mail: rkakc_s@caluniv.ac.in, kundu.raka@gmail.com).
**A. Chakrabarti** is an Associate Professor and Coordinator at the A. K. Choudhury School of Information Technology and Principal Investigator at CoE in Systems Biology and Biomedical Engineering, University of Calcutta, Kolkata, 700009, India (e-mail: acakcs@caluniv.ac.in, achakra12@yahoo.com)

**P. Lenka** is with Research and Development program at National Institute for the Orthopaedically Handicapped (NIOH), Kolkata, 700090, India (e-mail: lenka pk@yahoo.co.uk)